\documentclass{article}

\usepackage{microtype}
\usepackage{graphicx}
\usepackage{subfigure}
\usepackage{booktabs} 
\usepackage{tabularx}
\usepackage{makecell}
\usepackage{placeins}

\usepackage{bm}
\usepackage[fixamsmath]{mathtools}
  \mathtoolsset{showmanualtags,mathic,centercolon}
  
\usepackage{rotating}

\PassOptionsToPackage{hyphens}{url}\usepackage{hyperref}

\usepackage[accepted]{icml2021}

\icmltitlerunning{Mixed Effects Random Forests for Personalised Predictions of Clinical Depression Severity}

\begin{document}

\twocolumn[
\icmltitle{Mixed Effects Random Forests for Personalised Predictions of Clinical Depression Severity}
\vspace{-4mm}


\icmlsetsymbol{equal}{*}

\begin{icmlauthorlist}
\icmlauthor{Robert A. Lewis}{mit}
\icmlauthor{Asma Ghandeharioun}{equal,mit}
\icmlauthor{Szymon Fedor}{equal,mit}
\icmlauthor{Paola Pedrelli}{mgh}
\icmlauthor{Rosalind Picard}{mit}
\icmlauthor{David Mischoulon}{mgh}
\end{icmlauthorlist}

\icmlaffiliation{mit}{MIT Media Lab, Massachusetts Institute of Technology, Cambridge, MA, USA.}
\icmlaffiliation{mgh}{The Depression Clinical and Research Program, Massachusetts General Hospital, Boston, MA, USA}

\icmlcorrespondingauthor{Robert Lewis}{roblewis@media.mit.edu}

\icmlkeywords{Machine Learning, Mixed-Effects, Mental Health, Depression}

\vskip 0.15in
]



\printAffiliationsAndNotice{\icmlEqualContribution} 

\begin{abstract}
This work demonstrates how mixed effects random forests enable accurate predictions of depression severity using multimodal physiological and digital activity data collected from an 8-week study involving 31 patients with major depressive disorder. We show that mixed effects random forests outperform standard random forests and personal average baselines when predicting clinical Hamilton Depression Rating Scale scores (HDRS\textsubscript{17}). Compared to the latter baseline, accuracy is significantly improved for each patient by an average of 0.199-0.276 in terms of mean absolute error ($p \ll 0.05$). This is noteworthy as these simple baselines frequently outperform machine learning methods in mental health prediction tasks. We suggest that this improved performance results from the ability of the mixed effects random forest to personalise model parameters to individuals in the dataset. However, we find that these improvements pertain exclusively to scenarios where labelled patient data are available to the model at training time. Investigating methods that improve accuracy when generalising to new patients is left as important future work.
\end{abstract}

\vspace{-5mm}
\section{Introduction}

In recent years, smartphones and wearable technologies have become increasingly ubiquitous, while data processing and machine learning (ML) capabilities have matured. As a result, excitement has risen about a new digital era for mental health diagnosis, treatment, and prevention, where technology will be used to augment clinical workflows and to improve direct-to-consumer products such as digital therapeutics \cite{Doraiswamy2019,Hsin2018}. 

\textit{Digital phenotyping} refers to the collection of continuous, multimodal, and in-situ data from patients using smartphone and wearable devices \cite{Torous2016,Onnela2016}. It is considered central to digital mental health, as one can use ML on these data to enable the accurate forecasting and detailed understanding of mental health psychopathology (e.g., by discovering phenotypes that associate with mental health states). To give an example, a regressor could be learned to map clinical depression scores, such as the Hamilton Depression Rating Scale, HDRS \cite{hamilton1960rating}, onto these digital data streams. Given such data can be collected passively and continuously, more frequent predictions of depression severity (via the HDRS score) can be made. This can provide substantial improvements to the patient’s care: for example, by enabling the early detection of treatment response or acute phase onset (e.g., relapse), care pathways can be adjusted expeditiously to improve patient outcomes \cite{Huckvale2019}.

However, enabling this vision is not trivial due to a multitude of factors, including the high degrees of heterogeneity both in the presentation of mental health conditions and in patients’ digital data streams. In such data scenarios, where multiple observations exist for each patient (i.e., \textit{repeated measures} data) and these observations are not \textit{independent and identically distributed} across patients (i.e., the data are \textit{non-IID}), it becomes increasingly difficult to learn a single predictor that is accurate for all the patients in the cohort. Indeed, this performance degradation can be so severe that it has been observed that simple personal baselines -- such as calculating the mean or median of that target variable over observations in the training set -- often outperform state-of-the-art ML models \cite{Demasi2017MeaninglessCL,Pedrelli2020}. Such limitations are a significant barrier to using ML to augment mental health care and understanding. 

In this work, we directly address the question of the utility of ML for HDRS depression severity prediction by comparing mixed effects random forests to simpler baselines. Mixed effects methods extend standard ML models to handle heterogeneous data. They do so by learning a subset of parameters for each individual: as such, they can be considered a form of model \textit{personalisation}. Our contributions are twofold: first, we outline how we apply this technique to a \textit{digital phenotyping} dataset that was collected in a clinical context and contains features related to known biomarkers of depression (e.g., quality of sleep). Second, we report empirical results that suggest mixed effects methods significantly improve depression severity predictions over standard random forests and average personal baselines. 

\vspace{-2mm}
\section{Related Work}

\subsection{Machine Learning for Mental Health Prediction}
\label{sec:ml4dp}

There is promising evidence on the feasibility of using ML to predict mental health states (e.g., clinical scores of depression) from physiological and behavioural data. Features have been engineered from various modalities, including self-report surveys, location patterns, smartphone usage (e.g., social media), electrodermal activity, accelerometer, and heart rate data \cite{grunerbl2014smartphone, DeChoudhury2014PPD, Saeb2015MobileSensorCorrelates, canzian2015trajectories, suhara2017deepmood, Ghandeharioun2017, Pedrelli2020}. However, much work remains. For example, more longitudinal studies are required to collect multimodal data streams and annotate them with clinical scores so high-quality data is available to train ML models, and more quantitative contributions are required to identify ML methods that reliably outperform personal baselines. 

\subsection{Mixed Effects Machine Learning}

Mixed effects models are used in statistics and econometrics for longitudinal data, where \textit{repeated measures} are collected from individuals in the system \cite{Wu2006NonparametricRM,fitzmaurice2012applied}. They incorporate \textit{random effect} parameters into models in addition to the \textit{fixed effect} terms, which adjusts the model's assumptions to account for heterogeneous data with multiple sources of random variability (e.g., both intra- and inter-individual). As a result, mixed effects methods allow stronger statistical conclusions to be made about the factors that correlate with the observed variance. 

More recently, mixed effects methods have been considered in the context of ML \cite{Hajjem2011,Sela2012,NGUFOR2019}, where they are included as a way to improve predictive accuracy. By granting the model the flexibility to learn some \textit{random effect} parameters for each individual, predictions are \textit{personalised} and, thus, accuracy is increased. This improvement has been noted in several contexts though, to the best of our knowledge, prior work has not yet studied if mixed effects machine learning methods can improve accuracy when predicting mental health severity using multimodal digital phenotyping data. 

\section{Methods}
\label{sec:methods}

In this work we empirically assess the performance of a mixed effects random forest (MERF) method\footnote{We explicitly acknowledge the authors of the MERF theory \cite{Hajjem2014} and of the opensource Python implementation \cite{website:merf} whose work we build upon in this empirical paper.} on repeated measures HDRS scores. The method is referred to as \textit{mixed effects} as it contains both \textit{fixed effect} parameters -- i.e., those that are shared by all \textit{clusters}\footnote{In this context, clusters refer to the individuals in the system that generate repeated measures data (e.g., patients) and they are specified \textit{a priori} (i.e., it is assumed that the data from each patient forms a cluster and this does not change as the model is fit).} in the dataset -- and \textit{random effect} parameters -- i.e., those that are unique for each \textit{cluster}. Beyond the random effect parameters, we are interested in the random forest component of this method (the \textit{fixed effect}), given the random forest’s ability to maintain performance when there are many more features than observations (i.e., in the $p \gg n$ context) \cite{CHEN2012323}. The MERF model \cite{Hajjem2014} is defined by: 

\begin{align}
    \boldsymbol{Y_i} &= f(\boldsymbol{X_i}) + \boldsymbol{Z_ib_i} + \boldsymbol{\epsilon_i} \label{eq:nlme} \\ 
    \boldsymbol{b_i} &\sim N(0,\boldsymbol{D}) \label{eq:nlme_b} \\ 
    \boldsymbol{\epsilon_i} &\sim N(0,\boldsymbol{R_i}) \label{eq:nlme_e} \\ 
    \boldsymbol{V_i} &= \textrm{Cov}(\boldsymbol{Y_i}) = \boldsymbol{Z_i}\boldsymbol{D}\boldsymbol{Z_i^T} + \boldsymbol{R_i} \label{eq:nlme_ycov} \\
    \boldsymbol{R_i} &= \boldsymbol{\sigma^2I_{n_i}} \label{eq:nlme_var} 
\end{align}
\vspace{-2em}

Where $i {=} 1, ..., m$ are \textit{clusters} (i.e., patients) with $n_i$ observations each ($j {=} 1, ..., n_i$); $\boldsymbol{Y_i}$ is the regression target variable ($n_i{\times}1$); $\boldsymbol{X_i}$ is a design matrix of input features ($n_i {\times} p$) and $f(\boldsymbol{X_i})$ is the \textit{fixed effect} random forest estimator; $\boldsymbol{Z_i}$ is also a design matrix ($n_i {\times} q$), that usually contains a subset of features from $\boldsymbol{X_i}$; $\boldsymbol{b_i}$ are \textit{random effect} parameters ($q {\times} 1$) for each $i$, and $\boldsymbol{Z_ib_i}$ is assumed to be linear; $\boldsymbol{\epsilon_i}$ is the measurement error for each $i$; and, $\boldsymbol{D}$, $\boldsymbol{R_i}$ and $\boldsymbol{V_i}$ are covariance matrices with assumptions that i) $\boldsymbol{b_i}$ and $\boldsymbol{\epsilon_i}$ are independent of each other, and ii) between-cluster variation is the only source of correlation between the repeated measurements $\boldsymbol{Y_i}$, such that intra-subject measurement errors $\boldsymbol{\epsilon_i}$ are independent and $\boldsymbol{R_i}$ is diagonal per \eqref{eq:nlme_var}.

While $\boldsymbol{Z_i}$ may include many features, only a random intercept is used in the experiments of this paper. Thus, $\boldsymbol{Z_i}$ becomes a ($n_i {\times} 1$) vector of ones, and so (\ref{eq:nlme}-\ref{eq:nlme_var}) can be expressed for each observation $ij$ as: 

\vspace{-2em}
\begin{align}
    Y_{ij} &= f(X_{ij}) + b_i + \epsilon_{ij} \label{eq:intercept_only} \\ 
    E(Y_{ij}|b_i) &= f(X_{ij}) + b_i \label{eq:cond_exp} \\  
    E(Y_{ij}) &= f(X_{ij}) \label{eq:uncond_exp}
\end{align}
\vspace{-2em}

$E(Y_{ij}|b_i)$ is the \textit{conditional expectation} of the model, and can be computed for clusters $i$ that are known when the model is fit. $E(Y_{ij})$ is the \textit{unconditional expectation} and it represents how the model predicts for a cluster $i$ that is unknown at training time. These expectations are useful for interpreting model prediction errors for existing versus new users. The model parameters are fit using an expectation maximisation (EM) procedure with convergence monitored by a generalised log likelihood objective function. Further details of this training algorithm can be found in Supplementary Material (SM) Section \ref{sec:sup_methods}.

\section{Experimental Work}

\subsection{Data}

The dataset consists of 1,643 days of data collected from 31 patients with major depressive disorder (MDD), where MDD is categorised per the DSM-IV\footnote{The Diagnostic and Statistical Manual of Mental Disorders 4\textsuperscript{th} edition (DSM-IV) provides a taxonomy for the classification of mental disorders \cite{dsm}. It is published by the American Psychiatric Association and is used in clinical practice in the USA.} and patients have a score of $\geq19$ on the 28-item Hamilton Depression Rating Scale, HDRS\textsubscript{28} \cite{hamilton1960rating}. Patients were monitored for 8 weeks and several categories of data were collected, with multiple observations per participant (i.e., the data has repeated measures). First, clinical assessments were performed by clinicians during 6 visits (once during screening followed by 5 bi-weekly visits during the 8-week monitoring period). These clinical scores include the HDRS\textsubscript{28}, as well as a shorter 17-item HDRS scale, HDRS\textsubscript{17}, which is commonly used to measure depressive symptom severity in clinical trials. Second, multimodal data was collected passively and continuously using mobile phones \cite{website:movisens} and physiological-sensor wristbands \cite{website:empatica}. This data includes modalities known to associate with depressive symptomatology, including electrodermal activity (EDA) and heart rate variability (HRV), sleep characteristics, physical activity, digital activity (e.g., location and smartphone usage), and weather information. Features are created at various levels of temporal aggregation, including hourly and daily. Given the limited space, the feature design is discussed at length in Table~\ref{tab:datasummary} of SM Section~\ref{sec:sup_data}, as well as in our previous work \cite{Pedrelli2020}. 

For the experiments, the dataset contains 2,820 features, and its rows are filtered to only include data captured on days with clinical scores, resulting in 149 observations in total\footnote{While 5 clinical scores are expected per participant during the observation period (excluding the initial screening visit), several participants had missing observations (6 are missing in total).}. Within observations, subsets of features are sometimes missing as a result of various events in the study period (e.g., sensor not worn for part of the day). These missing values are set to -1 before the model is trained.

\vspace{-2mm}
\subsection{Evaluation Scenarios}
\label{sec:scenarios}

Three evaluation scenarios are considered. First, a \textit{Random Split} setting with a train:test split of 70:30. Second, a \textit{Time Split} setting, where the initial 3 observations per patient are used for training, and the remaining observations are used for testing. In practice, this scenario reflects making further predictions for existing patients (i.e., those that have already received clinical assessments). Finally, a \textit{User Split} setting is considered, where observations for one patient are held out from the model as a testing set, while the observations for the remaining patients are used in the training set. The sampling is then repeated 31 times so all patients are in the testing set exactly once. In practice, this scenario reflects making predictions for new patients.

\begin{figure}
    \begin{center}
    \centerline{\includegraphics[width=\columnwidth]{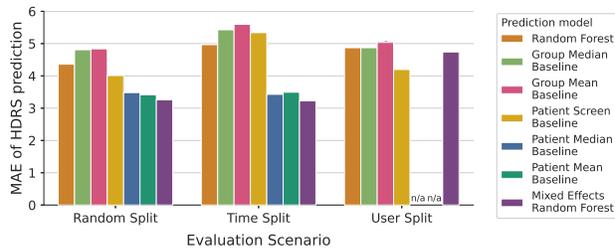}}
    \caption{Mean absolute error (MAE) of testing set HDRS\textsubscript{17} prediction by evaluation scenario and model type. The MAE is reported here at the \textit{group-level}: i.e., it is the average error across all observations in the testing set. The scenarios and models are described in Sections~\ref{sec:scenarios} \& \ref{sec:models}, respectively.}
    \label{fig:maebar}
    \end{center}
    \vspace{-10mm}
\end{figure}

\begin{table*}[!t]
    \caption{\textit{Participant-level} errors by scenario. The average of the mean absolute errors (MAE) for each participant is reported (Avg. Err.). The \textit{user lift} metric represents the average performance increase of the \textit{Mixed Effects Random Forest} (MERF) compared to the most accurate personal baseline (PBL). PBL is: the \textit{Patient Mean} if \textit{Random} scenario; the \textit{Patient Median} if \textit{Time Split} scenario; else, the \textit{Patient Screen} score if \textit{User Split} scenario. A one-sample one-tailed nonparametric permutation test ($p\leq0.05$) is used to test if the \textit{user lift} value is significantly greater than zero. The p-values from this test are aggregated over independent experiments (with $N{=}10$ random seed repeats) using Fisher’s method. The worst-case error metric (WC Err.) representing the worst error for any participant is also reported (see Section~\ref{sec:settings}). \textbf{NB}: the permutation test is not calculated in the \textit{User Split} scenario as the \textit{user lift} is clearly less than zero.}
    \label{tab:user_lift}
    \vskip 0.15in
    \begin{center}
    \begin{scriptsize}
    \begin{sc}
        \begin{tabular}{lr|rrrr|rr}
        \toprule
                Scenario &  N. Seeds &  Avg. PBL Err. &  Avg. MERF Err. &  Avg. User Lift &  User Lift p-value &  WC PBL Err. &  WC MERF Err. \\
        \midrule
                Random Split &        10 &          3.349 &           3.165 &           0.199 ($\uparrow$) &              \textbf{0.000} &        8.167 &         7.933 \\
                Time Split &        10 &          3.450 &           3.174 &           0.276 ($\uparrow$) &              \textbf{0.004} &        8.500 &         7.657 \\
                User Split &        10 &          4.198 &           4.739 &          -0.541 ($\downarrow$) &              n/a &       11.000 &        12.417 \\
        \bottomrule
        \end{tabular}
    \end{sc}
    \end{scriptsize}
    \end{center}
    \vskip -0.2in
\end{table*}

\vspace{-2mm}
\subsection{Models Assessed}
\label{sec:models}

Several simple baselines are implemented: the median and the mean HDRS\textsubscript{17} score of patients in the training set (\textit{Group Median} and \textit{Group Mean}, respectively); the HDRS\textsubscript{17} score of the patient at the initial screening visit (\textit{Patient Screen}); and, the median and mean HDRS\textsubscript{17} score of the patient from the training set observations (\textit{Patient Median} and \textit{Patient Mean}, respectively). Furthermore, to compare the mixed effects model to a standard ML approach, a random forest regressor without mixed effects is implemented (\textit{Random Forest})\footnote{The accuracies of other standard ML models are presented in SM Section~\ref{sec:sup_results}. These perform worse than the random forest and are excluded from the main body in the interest of brevity.}. Finally, the mixed effects random forest regressor is implemented per (\ref{eq:nlme}-\ref{eq:nlme_var}) with a random effect term for the intercept (\textit{Mixed Effects Random Forest; MERF}). In all cases, the raw HDRS\textsubscript{17} score is predicted (range 0-52) without any pre-processing adjustments.

\vspace{-2mm}
\subsection{Evaluation Metrics and Settings}
\label{sec:settings}

The mean absolute error (MAE) is used to evaluate accuracy. It is calculated at two levels. First, in Figure~\ref{fig:maebar} a single MAE error is calculated across all observations at the \textit{group-level}. Second, in Table~\ref{tab:user_lift} a MAE error is calculated for each participant (i.e., the mean error using only predictions from that participant). This second level deepens the assessment of MERF versus the personal baselines. Informed by suggestions from the literature \cite{Demasi2017MeaninglessCL}, the MAE at the \textit{participant-level} are used to derive a \textit{user lift} metric, which represents the improvement of the MERF model over the baseline (e.g., if the baseline MAE is 4 and the MERF MAE is 3, then the \textit{user lift} is 1). A corollary of this \textit{participant-level} approach is that one can formally test if, on average, the \textit{user lift} is significantly greater than zero. To do so a one-sample one-tailed nonparametric permutation test is performed, from which the p-values are reported in Table~\ref{tab:user_lift}. Finally, the \textit{worst-case error} is also reported, which represents the worst MAE for any participant, and is thus a measure of model robustness in this respect. Experimental settings are discussed further in SM Section \ref{sec:sup_settings}. 

\vspace{-2mm}
\subsection{Results}

Figure~\ref{fig:maebar} shows the MAE at the \textit{group-level}. In two scenarios -- the \textit{Random Split} and \textit{Time Split} -- the mixed effects random forest shows an improvement over both the standard random forest and, more importantly, the patient median and mean baselines. However, it is also noteworthy that the mixed effects approach provides less of an improvement in the \textit{User Split} scenario, and indeed using the patient HDRS\textsubscript{17} score at screening is a far better predictor in this setting.

Table~\ref{tab:user_lift} shows \textit{participant-level} errors and \textit{user lift} metrics. Improvement over the personal baselines is reflected at this level in the \textit{Random Split} and \textit{Time Split} scenarios, with the permutation tests suggesting the improvement is significant ($p\leq0.05$). Moreover, the worst-case error is also slightly improved in these scenarios by MERF, suggesting it also helps to improve prediction robustness across individuals.

\vspace{-2mm}
\section{Discussion}
\vspace{-1mm}

These results suggest that a mixed effects approach allows random forests to significantly outperform baselines in HDRS\textsubscript{17} predictions when patients in the testing set have also contributed some data to the training set. Compared to the standard random forest, this benefit likely stems from the ability of the mixed effects model to fit a random effect intercept parameter for each patient ($\boldsymbol{b_i}$). Such flexibility ensures the predictions of the learned model are not centred around a group average HDRS\textsubscript{17} score, but rather are adjusted by the average observed scores for each patient (cf. the conditional expectation \eqref{eq:cond_exp}). That said, given the significant lift of MERF over personal average baselines, it is clear that the model learns more than just a \textit{participant-level} intercept. Indeed, it is probable that this additional lift is due to the random forest fixed effect terms, $f(\boldsymbol{X_i})$, that -- when estimated using the EM procedure (cf. SM Section~\ref{sec:sup_methods}) -- learn relations between the multimodal input features and the HDRS\textsubscript{17} scores (net of patient specific random effect values) that can be shared across patients. 

Nevertheless, this method has limitations. Most notably, it does not provide a lift in the \textit{User Split} scenario. In contrast to the \textit{Time Split}, in the \textit{User Split} scenario no random intercept parameter can be learned for the target user, and thus the model can only predict using the fixed effect parameters learned from the patients in the training set (cf. the unconditional expectation \eqref{eq:uncond_exp}). As such, its accuracy is similar to that of a standard random forest in this scenario. 

As future work we will first introduce additional random effect parameters to assess if this further improves accuracy in the \textit{Random Split} and \textit{Time Split} scenarios. Second, we intend to further analyse the model's errors on a patient-level, assessing if patient characteristics (e.g., the properties of their training data distribution) correlate with them. Third, we intend to repeat these analyses with alternative target variables and datasets (e.g., including individuals without a current MDD diagnosis) to further understand how this method generalises to related mental health prediction tasks. Fourth, as we do not currently use all days of data collected (i.e., days without clinical scores are excluded), we will consider ways to incorporate this information using time-lagged approaches. Fifth, we will attempt to improve model generalisation to new patients (i.e., the \textit{User Split} scenario), e.g., by using patient characteristics to compute initial random effect parameter values for the new patients.

Finally, it is important to comment on the clinical significance of these results. While a lower MAE would be required to enable precision interventions, e.g., just-in-time-adaptive interventions \cite{NahumShani2018JustinTimeAI}, an MAE $\approx \! 3.2$ across patients is still useful for patient monitoring. For example, per the HDRS\textsubscript{17}, if a patient’s score worsens from 0 (a level indicating recovery) to $\geq \!15$ (indicating relapse), then even with an MAE $\approx \! 3.2$ a relapse prediction can still be made with reasonable confidence and used to alert the patient’s care team. 

\vspace{-2mm}
\section{Conclusion}
\vspace{-1mm}

This work has shown that extending random forests with random effect intercept parameters significantly improves accuracy over personal baselines when predicting clinical HDRS\textsubscript{17} depression scores. These findings only apply to scenarios where patients have labelled data in the training set, but nevertheless are a noteworthy contribution to improving the utility of ML methods for digital mental health.

\FloatBarrier

\typeout{}
\bibliography{example_paper.bib}
\bibliographystyle{icml2021}

\clearpage 

\appendix
\twocolumn[
\icmltitle{Supplementary Material: Mixed Effects Machine Learning for Personalised Predictions of Clinical Depression Severity}
]

\section{Mixed Effects Random Forest Expectation Maximisation Training Procedure}
\label{sec:sup_methods}

This section resumes the theoretical overview of the mixed effects random forest model introduced in Section \ref{sec:methods}. To estimate the unknown parameters of $f(\boldsymbol{X_i})$ and $\boldsymbol{b_i}$ in (\ref{eq:nlme}-\ref{eq:nlme_var}), maximum likelihood is used. Given a closed-form solution for parameter estimation does not exist, an expectation maximisation (EM) procedure is used to iteratively update the parameters to maximise a generalised log likelihood objective function (GLL). Given this estimation approach is not original to our work, we state the GLL without proof in (\ref{eq:pop_err}-\ref{eq:gll}). Likewise, we outline the EM procedure with the parameter update equations in Algorithm~\ref{alg:merf}. For a rigorous theoretical overview, we refer the reader to a reference on the theory of mixed effects parameter estimation \cite{Wu2006NonparametricRM} as well as to the work of the originators of the mixed effects random forest (MERF) approach \cite{Hajjem2011,Hajjem2014}. 

\vspace{-2em}
\begin{align}
    \boldsymbol{\epsilon_i}    &= [\boldsymbol{Y_i} - f(\boldsymbol{X_i}) - \boldsymbol{Z_ib_i}] \label{eq:pop_err} \\ 
    GLL(f, \boldsymbol{b_i}|\boldsymbol{Y}) &= \sum_{i=1}^n\{{\boldsymbol{\epsilon_i^TR_i^{-1}\epsilon_i}} + \boldsymbol{b_i^TD^{-1}b_i} \label{eq:gll} \\ 
                  & \quad\qquad + log|\boldsymbol{D}| + log|\boldsymbol{R_i}| \} \nonumber
\end{align}
\vspace{-2em}

\begin{algorithm}
  \small
  \caption{Fit Mixed Effects Random Forest with EM}
  \label{alg:merf}
\begin{algorithmic}
  \vspace{1mm}
  \STATE {\bfseries System:} MERF model defined by (\ref{eq:nlme}-\ref{eq:nlme_var})
  \vspace{1mm}
  \FOR{$r=0$ {\bfseries to} $N_{iterations}$}
        \vspace{1mm}
        \STATE {\textbf{E-step}:}
            \vspace{0.1em}
            \STATE {(i) Let $\boldsymbol{Y_i^*} = \boldsymbol{Y_i} - \boldsymbol{Z_i\hat{b}_{i(r-1)}}$} 
            \STATE {(ii) Fit random forest to $\boldsymbol{Y_i^*}$ to obtain $\hat{f}(\boldsymbol{X_i})_{(r)}$} 
            \STATE {(iii) Fit $\boldsymbol{\hat{b}_{i(r)}} = \boldsymbol{\hat{D}_{(r-1)}Z_i^T\hat{\boldsymbol{V}}_{i(r-1)}^{-1}}(\boldsymbol{Y_i} {-} \hat{f}(\boldsymbol{X_i)_{(r)}})$} 
            \vspace{0.2em}
            \STATE {Where: $\boldsymbol{\hat{V}_{i(r-1)}}$ is calculated by $\boldsymbol{\hat{D}_{(r-1)}}$ \& $\boldsymbol{\hat{R_i}_{(r-1)}}$ in \eqref{eq:nlme_ycov}}
        \vspace{-2mm}
        \STATE {\textbf{M-step}:} 
            \vspace{0.1em}
            \STATE {(i) Update: $\boldsymbol{\hat{\sigma}_{(r)}^2} = N^{-1} \sum_{i=1}^n\{{\boldsymbol{\hat{\epsilon}_{i(r)}^T\hat{\epsilon}_{i(r)}}} \quad + $}
            \vspace{0.2em}
            \STATE{\hspace{25mm} $ \boldsymbol{\hat{\sigma}_{(r-1)}^2}[n_i - \boldsymbol{\hat{\sigma}_{(r-1)}^2} \textrm{trace}(\boldsymbol{\hat{V}_{i(r-1)}})]\}$} 
            \vspace{0.4em}
            \STATE {\hspace{3mm} And: \hspace{2mm} $\boldsymbol{\hat{D}_{(r)}} = n^{-1} \sum_{i=1}^n\{{\boldsymbol{\hat{b}_{i(r)}\hat{b}_{i(r)}^T}} \quad + [\boldsymbol{\hat{D}_{(r-1)}} {-} $}
            \vspace{0.2em}
            \STATE{\hspace{25mm} $ \boldsymbol{\hat{D}_{(r-1)}Z_i^T \hat{V}_{i(r-1)}^{-1}Z_i\hat{D}_{(r-1)}}]\}$} 
            \vspace{0.2em}
            \STATE {\hspace{3mm} Where: $\boldsymbol{\hat{\epsilon}_{i(r)}} = \boldsymbol{\hat{Y}_i} - \hat{f}(\boldsymbol{X_i}) - \boldsymbol{Z_i\hat{b}_i}$ (per \eqref{eq:pop_err})}
        \vspace{1mm}
  \ENDFOR
\end{algorithmic}
\end{algorithm}

\section{Summary of the Multimodal Dataset with Clinical Annotations}
\label{sec:sup_data}

Table~\ref{tab:datasummary} summarises the data collected during the observational study and how it was engineered into features for the machine learning experiments. The feature designs were informed by prior work identifying biomarkers / correlates of depressive symptomatology. In total, the dataset contains 2,820 input features and a a 1-dimensional HDRS\textsubscript{17} as the target variable. 1,643 rows of data are collected in total (each representing one day of patient observations). However, in the experiments of this paper, the rows are filtered to only include days where a clinical HDRS\textsubscript{17} score was reported. Thus, the total number of observations in the assessed dataset is 149. This corresponds to 5 clinical measures for 31 patients, minus 6 observations which are missing for data quality reasons.

\begin{table}[!t]
    \caption{Data were collected from several modalities and engineered into features using various aggregation techniques. Each row in the dataset represents the observations for a given day. 1,643 days of data are collected in total. The final model uses 2,820 features as input and a 1-dimensional HDRS score variable as its target.}
    \label{tab:datasummary}
    \begin{center}
    \begin{scriptsize}
        \begin{tabularx}{0.48\textwidth} { 
             >{\raggedright\hsize=0.175\hsize}X 
            | >{\hsize=0.825\hsize}X  }
                \toprule
                \textbf{Modality}& \textbf{Description} \\ [0.3ex] 
                \midrule
                HDRS\textsubscript{17} clinical score (target) & The 17-item Hamilton Depression Rating Scale \cite{hamilton1960rating} is a clinician-administered depression assessment scale. It was administered 6 times during the study: once during the initial screening visit, once at the beginning of the 8-week observation period, and then every other week for the remainder of the study. \\ 
                & The regression target variable is created by summing the HDRS\textsubscript{17} items to create a total score with a range of 0 to 52. The mean of the HDRS\textsubscript{17} scores in this dataset is 17.7, the minimum is 5, and the max is 31. \\ \midrule \midrule
                Physiology: Electrodermal Activity (EDA) & Skin conductance level (SCL) and skin conductance response (SCR) are measured on the left and right wrists. Previous work has identified associations between skin conductance and stress / mental health \cite{Sano2018}. \\ 
                & Various aggregated statistics are calculated on SCL \& SCR (e.g., number of peaks and amplitude) at various temporal aggregations (e.g., hourly, daily, as well as aggregations to night, morning, etc.). Statistics are calculated for each wrist and for the difference between wrists. \\ \midrule
                Physiology: Heart Rate Variability & HRV is measured on the left and right wrists. HRV is often found to associate with Major Depressive Disorder \cite{koch2019}. \\
                (HRV) & Various HRV metrics are calculated in the time and frequency domains (e.g., AVNN, pNN50, rMSSD, SDANN, sDNNIDX, rrSDNN, PSD of the high-, low, very-low frequency signals, HF/LF ratio, etc.) at various temporal aggregations / periods in the day (e.g., hourly, daily, as well as aggregations to night, morning, etc.). \\
                & Average statistics for heart rate (HR) are also calculated on both wrists at various temporal aggregations in the day. \\ \midrule
                Sleep & Sleep time is calculated (over 24 hours and during the night). Other sleep characteristics are calculated using actigraphy, such as sleep onset time, wakeups, maximal night uninterrupted sleep, and a sleep regularity index \cite{Ghandeharioun2017}. Sleep disturbance is recognised as a core symptom of depression \cite{Nutt2008}. \\ \midrule
                Motion / Physical Activity & Features for motion frequency (i.e., fraction of time in motion within a period) and magnitude (i.e., the intensity of the motion) were calculated at various temporal aggregations using accelerometer data collected from the left and right wrists. \\ \midrule
                Digital Activity (smartphone) & An Android app \cite{website:movisens} was used to collect smartphone activity data, including streams for location, call and messaging (sms) activity, and app usage and screen on / off time. Average statistics (mean, counts, sums, variance, etc.) were calculated for these streams at various temporal aggregations. \\ \midrule
                Environment (weather) & Features related to the weather were also calculated (e.g., temperature, precipitation, humidity, UV index, etc.). The DarkSky API \cite{website:darksky} was used to obtain historical weather information for each participant by using their location collected through the MovisensXS app. \\ \bottomrule
                \hline 
        \end{tabularx}
    \end{scriptsize}
    \end{center}
\vspace{-5mm}
\end{table}

\section{Supplementary Information on the Experimental Settings}
\label{sec:sup_settings}

The experiments for each scenario were repeated 10 times with a different seed used for the random sampling scenario. This reduces the effect of data distribution and model training artefacts on the reported results. The hyperparameters of the standard random forest were tuned using a grid search, where values of the number of features, the max depth of trees, and the number of samples required to form splits and leaves are tuned. However, no hyperparameter tuning was performed on the mixed effects random forest (though this could be pursued as future work). 

A one-sample one-tailed nonparametric permutation test was implemented to formally test the significance of the patient-level user lift values \cite{Demasi2017MeaninglessCL}. A permutation test allows one to calculate a \textit{p-value} for a statistical test without requiring assumptions about the characteristics of the sampling distribution(s). It first calculates the test statistic on the observed sample(s). It then combines the samples, permutes them, and randomly samples from the combined sample. On each of these resamples, the test statistic is recalculated and compared to the observed test statistic (i.e., that from the original samples), allowing one to empirically calculate the probability -- i.e. the \textit{p-value} -- of observing test statistic values at least as extreme as the observed value. One can then use this \textit{p-value} with a pre-specified alpha (e.g., 0.05) to accept or reject the null hypothesis. To aggregate the \textit{p-values} in Table~\ref{tab:user_lift} across the independent experiment repeats (i.e., different seeds) Fisher’s method was used.

It is also worth noting why the one-tailed permutation test is not performed in the \textit{User Split} scenario in Table~\ref{tab:user_lift}. As the test is one-tailed, it only assesses if the \textit{user lift} is significantly greater than zero. For the \textit{User Split} scenario, the \textit{MERF} error is clearly worse than the \textit{Patient Screen} baseline and thus the \textit{user lift} is greater than zero. The one-tailed permutation test is thus not performed, as it does not test this side of the test statistic distribution, and thus reporting a non-significant \textit{p-value} here may seem incongruous with the magnitude of the negative user lift value.

The system is implemented in Python 3. Scikit-learn \cite{scikit-learn} is used for the machine learning models, with the mixed effects model also using the \textit{merf} Python library \cite{website:merf}. The permutation tests were implemented using \textit{MLxtend} \cite{raschkas_2018_mlxtend}.

\section{Supplementary Results: Comparison to Standard Machine Learning Baselines}
\label{sec:sup_results}

\begin{figure}[!t]
    \vskip 0.2in
    \begin{center}
    \centerline{\includegraphics[width=\columnwidth]{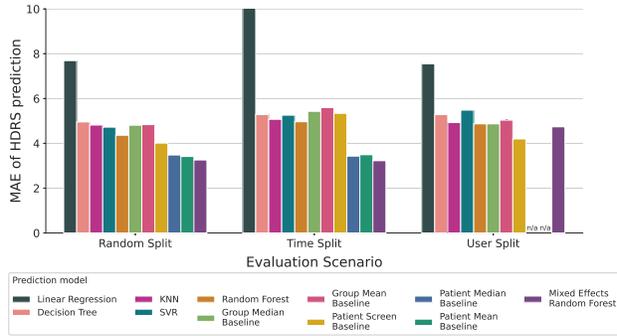}}
    \caption{Extended results set including standard machine learning models as additional baseline comparisons. Mean absolute error (MAE) of testing set HDRS\textsubscript{17} prediction by evaluation scenario and model type. The MAE is reported here at the \textit{group-level}: i.e., it is the average error across all observations in the testing set. The scenarios are described in Sections~\ref{sec:scenarios}, and the extended set of models are described in Section ~\ref{sec:sup_results}.}
    \label{fig:maebar_w_ml_baselines}
    \end{center}
    \vskip -0.2in
\end{figure}

To further evidence the benefits of the mixed effects random forest (MERF) model over standard baselines, Figure~\ref{fig:maebar_w_ml_baselines} displays an extended results set that includes the MAE of standard machine learning models, in addition to the mixed effects random forest and simple average baselines discussed in Section~\ref{sec:models}. The scenarios are consistent with those described in Section~\ref{sec:scenarios}. Given the performance of all of these machine learning models are similar to (or worse) than the group level averages, they are excluded from the main body in the interest of brevity. However, we discuss their details here for completeness. 

Hyperparameter tuning is performed on all of the machine learning models (with the exception of the mixed effects random forest) before the testing set MAE metric is generated. \textit{Linear Regression} refers to a standard regression model with linear parameters, and different hyperparameters are assessed for the strength and nature of regularisation (i.e., L1 vs L2). Its notably poor performance may result from either the data setting -- where there are many more features than observations (i.e., the $p \gg n$ context) -- or its inability to model nonlinear relations between features, or both. \textit{Decision Tree} refers to a standard decision tree regression, and hyperparameters of its max depth, as well as the number of samples required to form splits and leaves are tuned. \textit{KNN} refers to a standard k-nearest neighbours regression, and the number of neighbours optimised in the hyperparamter tuning. Finally, \textit{SVR} refers to a support vector regression model with a radial basis function kernel (RBF), and the kernel coefficient values (\textit{gamma}) and L2 regularisation strength (\textit{C}) are tuned. The models were implemented using Scikit-learn \cite{scikit-learn}.

\end{document}